\pdfoutput=1
\pdfoutput=1

\documentclass[11pt]{article}

\usepackage[final]{acl}

\usepackage{times}
\usepackage{latexsym}

\usepackage{times}
\usepackage{latexsym}
\usepackage{tcolorbox}
\usepackage{multirow}

\usepackage[T1]{fontenc}

\usepackage[utf8]{inputenc}

\usepackage{microtype}
\usepackage{array}
\usepackage{makecell}

\usepackage{inconsolata}

\usepackage{graphicx}
\usepackage{multirow}
\usepackage{bbding}
\usepackage{pifont}
\usepackage{subcaption}
\tcbuselibrary{breakable}
\tcbuselibrary{skins}
\usepackage{booktabs}
\usepackage{amsfonts}
\usepackage[ruled, vlined]{algorithm2e}

\usepackage[T1]{fontenc}

\usepackage[utf8]{inputenc}

\usepackage{microtype}

\usepackage{inconsolata}

\usepackage{graphicx}

\title{\includegraphics[height=1.5em]{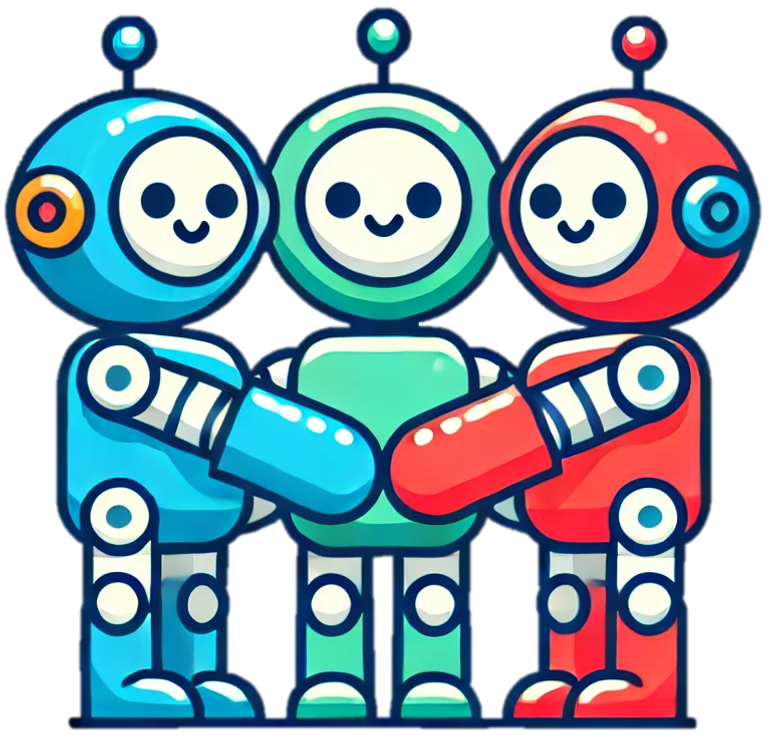} WorkTeam: Constructing Workflows from Natural Language with Multi-Agents}

\author{Hanchao Liu$^{\dagger}$ \and Rongjun Li$^{\dagger}$ \and Weimin Xiong$^{\ddagger}$ \and Ziyu Zhou$^{\dagger}$ \and Wei Peng$^{\dagger}$ \\
        $^{\dagger}$IT Innovation and Research Center, Huawei Technologies\\
        $^{\ddagger}$National Key Laboratory for Multimedia Information Processing,\\
        School of Computer Science, Peking University\\
        \texttt{\{liuhanchao2, lirongjun3, zhouziyu8, peng.wei1\}@huawei.com}\\
        \texttt{wmxiong@pku.edu.cn}}

\begin{document}
\maketitle
\begin{abstract}
Workflows play a crucial role in enhancing enterprise efficiency by orchestrating complex processes with multiple tools or components. However, hand-crafted workflow construction requires expert knowledge, presenting significant technical barriers. Recent advancements in Large Language Models (LLMs) have improved the generation of workflows from natural language instructions (aka NL2Workflow), yet existing single LLM agent-based methods face performance degradation on complex tasks due to the need for specialized knowledge and the strain of task-switching.
To tackle these challenges, we propose WorkTeam, a multi-agent NL2Workflow framework comprising a supervisor, orchestrator, and filler agent, each with distinct roles that collaboratively enhance the conversion process. As there are currently no publicly available NL2Workflow benchmarks, we also introduce the HW-NL2Workflow dataset, which includes 3,695 real-world business samples for training and evaluation. Experimental results show that our approach significantly increases the success rate of workflow construction, providing a novel and effective solution for enterprise NL2Workflow services.
\end{abstract}

\section{Introduction}

Workflows, comprising reusable processes that integrate multiple tools or components in a specific logic sequence, can significantly enhance enterprise efficiency~\cite{ayala2024reducing}. 
Traditional workflow construction methods require numerous manual steps to orchestrate components, demanding specialized expertise ~\citep{chi1981categorization, chi2014nature, faloughi2014simplean}.
In contrast, automated commercial systems can directly convert natural language instructions into workflows, offering a more convenient and technically accessible approach. 

\begin{figure}[h!]
    \centering
    \includegraphics[width=0.45\textwidth]{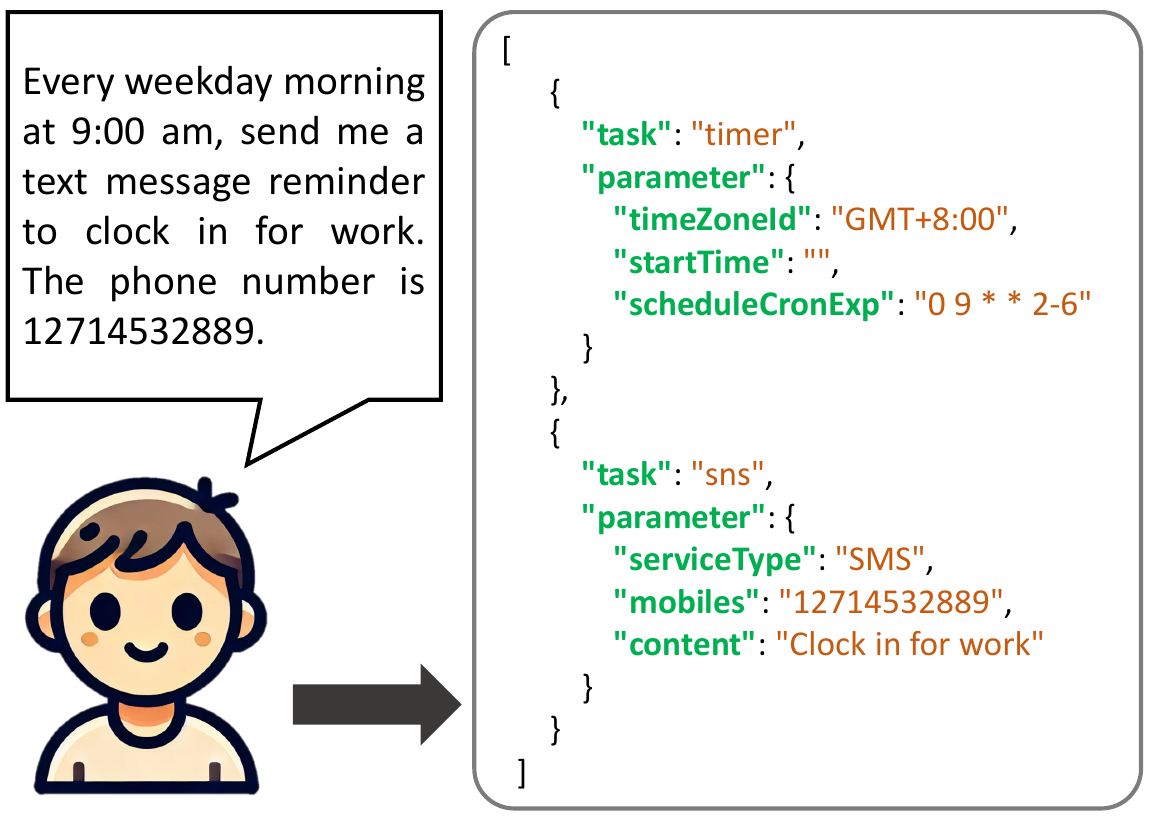}
    \caption{An example of generating workflows (JSON format) from text instruction.}
    \label{fig:NL2Workflow_example}
\end{figure}

With the rapid development of Large Language Models (LLMs)~\citep{achiam2023gpt, dubey2024llama} and LLM agents~\cite{xiong2024watch}, 
researchers have begun to utilize them as backbones to develop Natural Language to Workflows (NL2Workflow) systems.
\citet{zeng2023flowmind} directly prompted a LLM to generate workflows, while \citet{ayala2024reducing} improved this process by adopting a Retrieval-Augmented Generation~(RAG) approach to enhance the quality of the generated workflows.
Although they can produce workflows for simple scenarios, a significant gap remains compared to human performance in handling complex real-world instructions.


Crafting a workflow~(Figure~\ref{fig:NL2Workflow_example}) for real-life scenarios involves coordinating several tasks, from comprehending human intent, selecting appropriate components, to orchestrating the task flow and accurately configuring each component's parameters~\citep{wang2024agent}.
It's quite challenging to rely on a single LLM agent to handle the entire process, as different tasks may require specialized knowledge and skills. The need to switch between multiple tasks could potentially affect its performance on any individual task~\citep{gabriel2020artificial}.

To address this challenge, we draw inspiration from software development, where requires collaboration among multiple team members with diverse skill sets is essential~\citep{basili1989software, sawyer1998software}.
Specifically, we propose \textbf{WorkTeam}, a multi-agent framework that integrates multiple agents to collaboratively accomplish the NL2Workflow task.
WorkTeam consists of three agents with distinct roles: the supervisor, the orchestrator and the filler~(Figure~\ref{fig:overall_architecture}).
The supervisor agent is responsible for understanding the user's intent and coordinating the orchestrator agent and the filler agent. Upon receiving the user intent parsed by the supervisor agent, the orchestrator agent selects the appropriate components and arranges them into a suitable workflow schema. The filler agent then retrieves the documentation for relevant components and fills in accurate parameters, turning it into a fully operational workflow.
Our framework enables different agents to perform their respective tasks accurately and  communicate efficiently, thereby effectively constructing workflows.
Moreover, since no publicly available NL2Workflow benchmarks exist, we construct the \textbf{HW-NL2Workflow} dataset from real production scenarios, comprising 3,695 entries for training and evaluation.
Extensive experiments show that WorkTeam significantly improves workflow construction accuracy compared to existing methods, and further analysis validates the effectiveness of our framework.

Our contributions are summarized as follows:
\begin{itemize}
    \item For the first time, we introduced a multi-agent framework into the NL2Workflow task, effectively enhancing the automation of workflow construction.
    \item We construct the HW-NL2Workflow dataset, comprising 3,695 entries of real-world enterprise business data for training and evaluation.
    \item Extensive experimental results on HW-NL2Workflow demonstrate the superior performance of our method and the effectiveness of each framework component.
\end{itemize}

\section{Related Work}
\subsection{Natural Language to Workflow}

Recent advancements in LLMs have enabled the conversion of natural language instructions into logical outputs, such as code~\citep{xiong2023program, hong2024metagpt,jiang2024survey} and SQL~\citep{fu2023catsql,lian2024chatbi}, making it increasingly viable for commercial applications. 
Workflows, which serve as a structured form of task orchestration, automate repetitive activities across various industrial applications, such as data entry and invoice processing~\citep{villar2021robotic}.
To reduce technical barriers and expand commercial adoption, researchers are now focusing on generating workflows directly from natural language instructions.
For example, Microsoft~\cite{el2023automatic} and ServiceNow~\cite{gorrono2023generating} have patented systems that apply a machine learning model to transfer user-input text instructions into executable workflows. 
\citet{zeng2023flowmind} developed FlowMind, a system that employs LLMs to automatically generate workflows from user queries, enhancing automation in financial services while maintaining data security.
To improve the quality of generated workflows, \citet{ayala2024reducing} proposed an RAG-based method for NL2Workflow conversion. Upon receiving user instructions, their approach first retrieves relevant components and then generates workflows based on these components, effectively reducing hallucination issues. Although these methods have shown some success, single LLM-based approaches often suffer performance degradation in real-world commercial applications due to a lack of specialized knowledge and the strain of task-switching when handling complex instructions. 

\subsection{Multi-Agents}
Recently, LLM agents have been developed to understand and execute complex instructions, leading to improved interaction and more informed decision-making across various environments~\cite{xi2023rise,ruan2023tptu,wu2024chateda}. Along this line, multi-agent systems enhance functionality by utilizing the collective intelligence and specialized skills of multiple LLM agents, assigning distinct roles and facilitating interactions to better simulate complex real-world scenarios.

\begin{figure}[h!]
    \centering
    \includegraphics[width=0.48\textwidth]{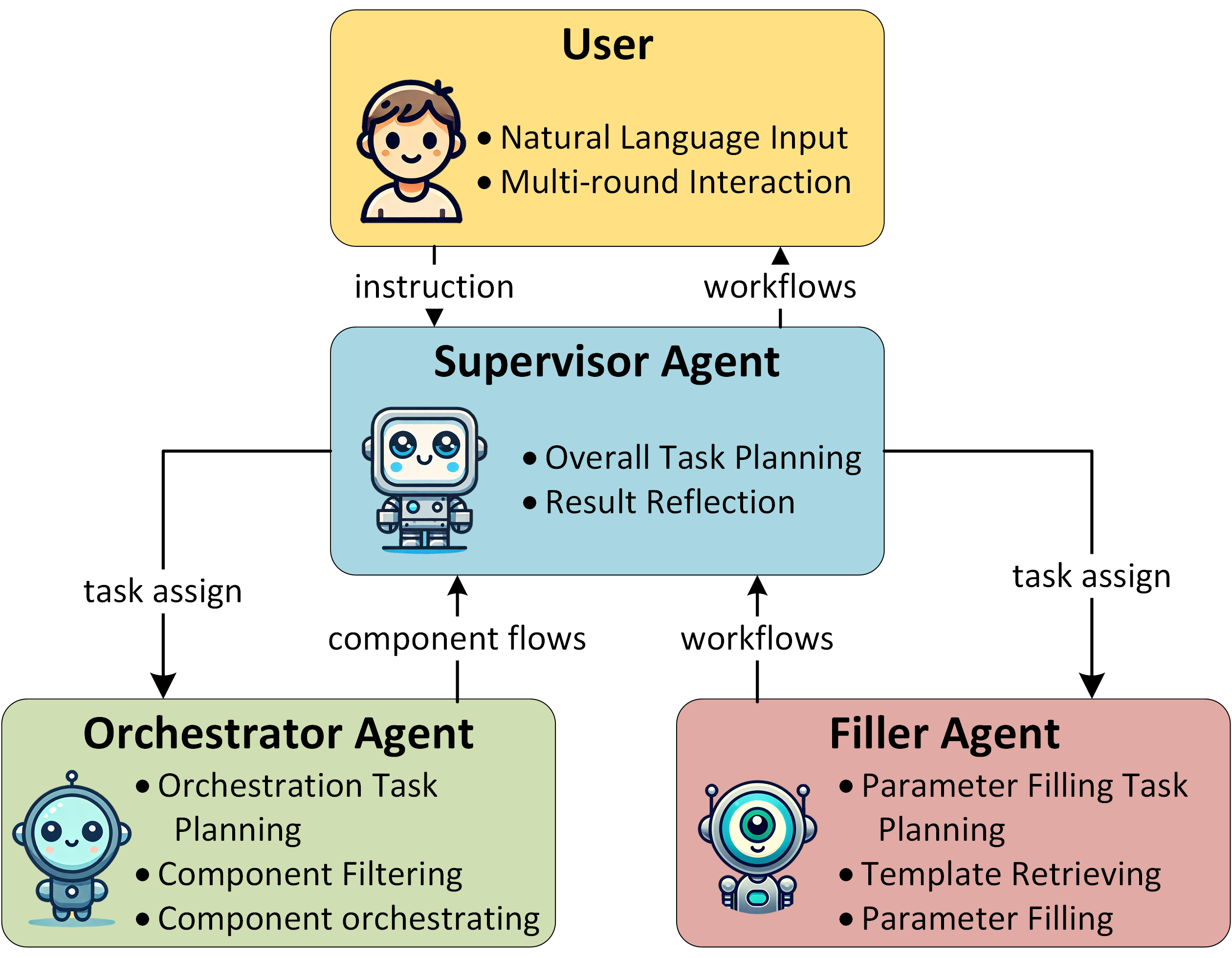}
    \caption{The overall architecture of the proposed WorkTeam framework.}
    \label{fig:overall_architecture}
\end{figure}

\citet{hong2024metagpt} introduced MetaGPT, a multi-agent collaborative framework for programming featuring six role-specific agents. This design, combined with Standardized Operating Procedures (SOPs), led to notable performance improvements in programming. In robotics, \citet{kannan2023smart} proposed SMART-LLM, a multi-agent framework for robot task planning. SMART-LLM decomposes user instructions into sub-tasks, assigns them to robots based on their skills, and coordinates execution to optimize task completion. In scientific experimentation, \citet{zheng2023chatgpt} implemented a multi-agent framework with agents specializing in areas like strategic planning, literature search, and coding. These agents collaborate with human researchers to improve the synthesis of complex materials. However, existing multi-agent approaches are generally designed for specific tasks and are not directly applicable to NL2Workflow.

In this paper, we propose a multi-agent approach to enhance NL2Workflow tasks, where agents with distinct roles and specialized skills collaborate to significantly boost workflow generation accuracy.


\section{Methods}

The WorkTeam framework comprises three agents: the supervisor agent, the orchestrator agent, and the filler agent. The overall structure of the framework is shown in Figure~\ref{fig:overall_architecture}. Upon receiving an end user's prompt, the supervisor agent initiates a task planning phase, decomposing the tasks into sub-tasks and invoking the orchestrator and filler agents in a coordinated manner to execute them. The orchestrator and filler agents handle component orchestration and parameter filling, respectively, using appropriate tools to complete these tasks. To further elucidate the functionality of WorkTeam, Figure~\ref{fig:workteam_example} in Appendix~\ref{sec:workteam_details} provides an operational example. The design and functionality of these agents are detailed as following.

\subsection{The Supervisor Agent}
The supervisor agent, as depicted in  Figure~\ref{fig:overall_architecture}, is responsible for two primary functions: task planning and result reflection. 
The task planning function allows the supervisor agent to dynamically plan based on user instructions.
For instance, when receiving a workflow creation instruction, the agent first calls the orchestrator agent for component orchestration, followed by the filler agent to populate the necessary parameters.
In contrast, for workflow modification instructions, the agent may invoke only the orchestrator or the filler agent.
This flexibility enables WorkTeam to efficiently execute user instructions.
Upon completion of task planning, the supervisor agent assigns tasks to either the orchestrator agent or the filler agent based on the planning results, to ensure the objectives are achieved.
After completing their tasks, the orchestrator and filler agents return the results to the supervisor agent for result reflection.
The next steps proceed only if the supervisor agent confirms the results are correct.
Otherwise, tasks are redirected to the appropriate agents for re-execution.

\subsection{The Orchestrator Agent}
The orchestrator agent selects appropriate components from the component set based on user instructions and arranges them in a logical order as implied by the instructions. To accomplish this, similar to the supervisor agent, the orchestrator agent first undertakes a dynamic planning process based on the input instructions $\mathrm{inst}_O$, which encompass user directives and, if available, feedback from the supervisor agent.
Subsequently, to ensure accurate orchestration results, the agent leverages two tools: the component filtering and the component orchestration tool, to finish the orchestration process based on the planning results. Next, we provide an overview of these two tools.

\paragraph{Component Filtering Tool} The primary objective of the component filtering tool is to select candidate components from the component set that are most relevant to the orchestrator agent's input instructions. These selected components serve as input for subsequent orchestration. Specifically, we use the SentenceBERT model~\cite{nils2019sentencebert} to extract embeddings for the orchestrator agent's input instructions $\mathrm{inst}_O$ and the descriptions $\mathrm{desc}_i$ for each component $t_i$, then compute the cosine similarity between the instruction and component embeddings to evaluate their relevance, as shown in Equations (\ref{cosine_score}) 
\begin{equation}
    s_i = \mathrm{Similarity}(\mathbf{e}_{inst}, \mathbf{e}_{desc}^i)
    \label{cosine_score}
\end{equation}
$\mathbf{e}_{inst}$ and $\mathbf{e}_{desc}^i$ represent their corresponding sentence embeddings for the input instructions and descriptions, $\mathrm{Similarity}$ is the cosine function, and $s_i$ is the similarity between $\mathbf{e}_{inst}$ and $\mathbf{e}_{desc}^i$.
Components with higher similarity scores are considered more relevant to the input instructions and prioritized as candidate components. We select the $top$-$k$ components based on descending similarity scores:
\begin{equation}
    C_{filtered} = \mathrm{TopK}(\langle t_1, s_1\rangle, \langle t_2, s_2\rangle, \dots, \langle t_n, s_n\rangle)
\end{equation}

\paragraph{Component Orchestration Tool} 
The primary objective of the component orchestration tool is to select and arrange a subset of components from the candidate components provided by the component filtering tool, based on the logic embedded in the input from the orchestrator agent, thereby generating a component flow. 
Given that the orchestration logic is embedded within the natural language instructions provided by the user, this process demands a high level of text comprehension. To address this challenge, we employ a large language model (LLM) as the component orchestration tool. The LLM can directly generate a component flow that incorporates the specified orchestration logic based on inputs of the orchestrator agent. The arranged component flow can be represented by:
\begin{equation}
    F_{C} = \mathrm{Tool}_{O}(\mathrm{inst}_O, C_{filtered})
\end{equation}
where $\mathrm{Tool}_{O}$ represents the component orchestration tool and $F_{C}$ is the generated component flow.

\subsection{The Filler Agent}

The filler agent populates parameters for each component in the given component flow $F_{C}$, transforming it into a complete workflow. Generally, the input of the filler agent $\mathrm{inst}_P$ comprises three main parts: the user textual instructions, the component flow provided by the orchestrator agent, and the feedback from the supervisor agent, with the latter two being optional. 
Similar to the supervisor agent and the orchestrator agent, the filler agent performs dynamic task planning upon receiving input. It decomposes the parameter filling task and then utilizes the template lookup tool and the parameter filling tool to ensure the accuracy and stability of the parameterization results. A detailed introduction to these two tools will be provided next.

\paragraph{Template Lookup Tool} The template lookup tool retrieves the parameter description $d_i$ and the blank parameter template $p_i$ associated with each component $t_i$ in $F_C$. The parameter description provides detailed information for each parameter, including its meaning, type, and allowable values. In contrast, the blank parameter template encompasses all parameters of the component, assigning a default value to each. By utilizing the pre-populated blank parameter template, only essential modifications to the component's parameters are required, significantly reducing the complexity of the parameter filling task.

\paragraph{Parameter Filling Tool} The parameter filling process begins once the tool has acquired three key elements: the orchestrated component flow $F_C$, the parameter description templates $d_i$ and the blank parameter templates $p_i$ for each component. With these in hand, the parameter filling tool's initial task is to analyze the input instructions,  extracting all relevant information necessary for accurate parameter instructions. Then, it need to populate the specified parameters in the blank templates based on their intended meanings, resulting in a complete workflow. Due to the complexity of this task, in this paper, we employ a LLM as the backbone for parameter filling tool. By providing the LLM with the input instructions $\mathrm{inst}_P$, component flow $F_C$, the looked-up parameter description templates $D=\{d_1, d_2, ..., d_m\}$, and the looked-up blank parameter templates $P=\{p_1, p_2, ..., p_m\}$ as prompts, the model is able to populate the parameters for each component in the stream, resulting in the generation of a complete workflow. The whole process can be represented by:
\begin{equation}
    F_{W} = \mathrm{Tool}_{P}(\mathrm{inst}_F, F_{C}, D, P)
\end{equation}
where $\mathrm{Tool}_{P}$ represents the parameter filling tool and $F_{W}$ is the generated workflow.

\section{HW-NL2Workflow}


Given the limited availability of publicly accessible datasets for NL2Workflow tasks and our focus on real-world commercial applications, we have developed HW-NL2Workflow, a novel dataset specifically designed to meet these needs. This dataset consists of 3,695 real-world enterprise workflows, making it suitable for both performance evaluation and tool training.


\subsection{Data Statistics}
\begin{table}[h]
\centering
\small
\begin{tabular}{l | c | c c c}
\toprule
\textbf{Split}           & \textbf{Type} & \textbf{Size}   & \textbf{\# Comp}       & \textbf{\# Param} \\ \midrule
\multirow{3}{*}{Train}   & Creation      & 2818            & 13993                   & 45696     \\
                         & Modification  & 562             & 2819                    & 9187      \\
                         & All           & 3380            & 16812                   & 54883     \\ \midrule
\multirow{3}{*}{Test}    & Creation      & 263             & 1269                    & 4244      \\
                         & Modification  & 52              & 252                     & 838       \\ 
                         & All           & 315             & 1521                    & 5082      \\ \bottomrule
\end{tabular}
\caption{Composition of HW-NL2Workflow. \# Comp and \# Param represent the number of components and parameters, respectively.}
\label{tab:data_statistics}
\end{table}

The HW-NL2Workflow dataset was created by collecting 3,695 workflows from our enterprise platform, each annotated by domain experts with natural language instructions. It is divided into training and testing sets, with detailed statistics provided in Table~\ref{tab:data_statistics}. Specifically, the dataset comprises 3,380 training samples and 315 testing samples.
On average, each workflow in the training set consists of 5.02 components, with each component having 3.26 parameters. In the testing set, workflows contain an average of 4.83 components and 3.34 parameters per component.
Additionally, the dataset encompasses both workflow creation and modification tasks, ensuring that WorkTeam can adapt to more flexible requirements.



\subsection{Component Resources}

In addition to data samples, the HW-NL2Workflow also provides comprehensive component resource information, including a component set $C$, a component parameter description set $T_{desc}$, and a blank parameter template set $T_{blank}$. These resource details provide sufficient component information to support workflow generation. Appendix~\ref{sec:component_resources} illustrates a few examples of the component resources of HW-NL2Workflow.

\subsection{Metrics}

We systematically evaluated the generated workflows from three perspectives:

\paragraph{Exact Match Rate (EMR)} Exact matching occurs when the generated workflow fully aligns with the ground truth, including both component sequence and parameter values. The exact match rate is calculated as $E_{acc} = N_{em} / N_{total}$, where $N_{em}$ and $N_{total}$ represent the exact matches and total test samples, respectively.

\paragraph{Arrangement Accuracy (AA)} Correct arrangement refers to the correctness of the sequence of components within the workflow generated by the model, irrespective of the correctness of the filled parameters. This metric primarily assesses the capability of the system to comprehend logical constructs in user instructions. Similarly, the arrangement accuracy is computed as $A_{acc} = N_{am} / N_{total}$, where $N_{am}$ represents the number of samples with accurate arrangement.

\paragraph{Parameter Accuracy (PA)} The parameter accuracy evaluates whether the parameters of the components in the generated workflow are consistent with those of the corresponding components in the ground truth. It is computed as $P_{acc} = N_{pm} / N_{p}$, where $N_{pm}$ and $N_{p}$ represent the number of matched parameters and the total number of parameters in the test set, respectively.

\section{Experiments}

\subsection{Configurations}

\paragraph{Model Configurations} WorkTeam is a multi-agent framework that supports implementation with various models. This subsection only focuses on the model configurations used in our experiments. All agents in our experiments are built on Qwen2.5-72B-Instruct~\cite{yang2024qwen2}. The prompt for all these agents are illustrated in Figure~\ref{fig:supervisor_prompt} to Figure~\ref{fig:filler_prompt} in Appendix~\ref{sec:workteam_details}.
The component orchestration tool and the parameter filling tool are implemented with LLaMA3-8B-Instruct~\cite{dubey2024llama}, fine-tuned on the HW-NL2Workflow dataset. 
Similarly, the component filtering tool is built using the SentenceBERT model, which has been fine-tuned with data from the HW-NL2Workflow dataset.

\paragraph{Training Data Configurations} 


The component filtering tool is built using the SentenceBERT model, trained with contrastive learning from paired text instructions and corresponding components.
The training data is directly derived from the HW-NL2Workflow dataset, with positive samples comprising text instructions and their relevant components, and negative samples comprising text instructions with unrelated components.

In our experiments, both the component orchestration and parameter filling tools are developed by finetuning a LLM.
The training data for the component orchestration tool includes the agent's input instruction, denoted as $\mathrm{inst}_O$, along with descriptions of the selected $top$-$k$ candidate components. The model's output is a workflow that consists solely of the names of these components.
For the parameter filling tool, the training data comprises the agent's input instruction $\mathrm{inst}_P$, the component flow $F_C$, the corresponding component parameter descriptions $D$, and blank parameter templates $P$, with the model's output being a complete workflow. 

\paragraph{Baselines}
Our experiments use a single LLM-based agent as the baseline, utilizing GPT-4o, Qwen2.5-72B-Instruct, Qwen2.5-7B-Instruct, and LLaMA3-8B-Instruct as backbone models.
These models generate workflows directly based on the input use instructions and in-context examples.
The prompts utilized for these approaches are detailed in Appendix~\ref{sec:baseline_details}. 
We also incorporate a RAG NL2Workflow method from ~\cite{ayala2024reducing} as an additional baseline. 
Due to the unavailability of the original source code, we implement our version using SentenceBERT as the retriever and LLaMA3-8B-Instruct as the generator, both trained on HW-NL2Workflow.


\subsection{Experiment Results}
\label{sec:experiment_results}

\begin{table}[h]
\centering
\resizebox{\linewidth}{!}{
\begin{tabular}{l | c c c}
\toprule
\textbf{Methods} & \textbf{EMR (\%)} & \textbf{AA (\%)} & \textbf{PA (\%)} \\ \midrule
GPT-4o                        & 18.1              & 71.4                   & 56.3               \\  
Qwen2.5-72B-Instruct          & 12.7              & 66.9                   & 51.5               \\  
Qwen2.5-7B-Instruct           & 3.5               & 25.4                   & 19.9               \\
LLaMA3-8B-Instruct            & 1.6               & 19.4                   & 16.6               \\
RAG~\citep{ayala2024reducing}  & 24.1              & 77.8                   & 60.3               \\ \midrule
\textbf{WorkTeam~(ours)}                          & \textbf{52.7}              & \textbf{88.9}                   & \textbf{73.2}               \\ \bottomrule
\end{tabular}
}
\caption{Comparison of experiment results of the baselines and our methods. }
\label{tab:result}
\end{table}

Table~\ref{tab:result} presents the performance comparison between WorkTeam and baseline methods on the HW-NL2Workflow test set. In our experiments, the single LLM agent approach generates workflows end-to-end by directly inputting all component information and user instructions. The prompts for this method are shown in Figure~\ref{fig:llm_prompt}. Table~\ref{tab:result} shows that the NL2Workflow task is highly challenging for single LLM-based method.
Top models like GPT-4o and Qwen2.5-72B-Instruct achieve only 18.1\% and 12.7\% EMR respectively, while smaller models such as Qwen2.5-7B-Instruct and LLaMA3-8B-Instruct are nearly ineffective, with EMRs of just 3.5\% and 1.6\%.
The RAG NL2Workflow method improves workflow construction accuracy compared to the single LLM agent approach, but EMR performance remains unsatisfactory.
In contrast, WorkTeam achieve an EMR of 52.7\%, an AA of 88.9\%, and a PA of 73.2\% on the HW-NL2Workflow test set, representing a comprehensive and significant improvement over baseline methods.

We attribute the performance enhancement of WorkTeam to task specialization and collaboration among multiple agents. 
The orchestrator and filler agents concentrate on their specific tasks, improving execution stability and accuracy, while the supervisor agent, responsible for task planning and result reflection, enhances robustness and flexibility.
Ablation studies, detailed in Table~\ref{tab:ablation_result}, further illustrate each agent's contribution.

\begin{table}[h]
\centering
\resizebox{0.9\linewidth}{!}{
\begin{tabular}{c c c | c c c}
\toprule
\makecell{\textbf{Supervisor} \\ \textbf{Agent}} & \makecell{\textbf{Orchestrator} \\ \textbf{Agent}} & \makecell{\textbf{Filler} \\ \textbf{Agent}} & \makecell{\textbf{EMR} \\ \textbf{(\%)}} & \makecell{\textbf{AA} \\ \textbf{(\%)}} & \makecell{\textbf{PA} \\ \textbf{(\%)}} \\ \midrule
\ding{51} & \ding{55} & \ding{55} & -                 & -                    & -                  \\
\ding{55} & \ding{51} & \ding{55} & -                 & 85.7                    & -                  \\
\ding{55} & \ding{55} & \ding{51} & -                 & -                      & -                  \\
\ding{55} & \ding{51} & \ding{51} & 49.8               & 85.7                   & 72.8                   \\ \midrule
\ding{51} & \ding{51} & \ding{51} & \textbf{52.7}              & \textbf{88.9}                   & \textbf{73.2}                \\ \bottomrule
\end{tabular}
}
\caption{Results of the ablation experiments for different agents.`-' represents the task cannot be completed.}
\label{tab:ablation_result}
\end{table}
The results in Table ~\ref{tab:ablation_result} demonstrates that both the orchestrator agent and the filler agent are essential for workflow generation, as the absence of either leads to task failure. Although the workflow can still be generated without the supervisor agent, the accuracy decreases from 52.7\% to 49.8\% compared to the complete WorkTeam. This indicates that the task planning and result reflection functions of the supervisor effectively facilitates collaboration between the orchestrator and filler agents, thereby enhancing workflow generation accuracy. 

To better illustrate the roles of WorkTeam's agents and its NL2Workflow process, we present a real-world case in Figure~\ref{fig:case study} of the Appendix~\ref{sec:case_study}. Additionally, we developed a commercial NL2Workflow system based on WorkTeam that effectively meets business requirements, as shown in Figure~\ref{fig:enterprise_workflow} of the same appendix.


\section{Conclusion}

In this paper, we present WorkTeam, a novel multi-agent framework designated to enhance workflow automation in enterprise environments. 
Three specialized agents — supervisor, orchestrator, and filler agents — collaborate to overcome the limitations of a traditional LLM agent-based method, resulting in substantial improvements to workflow generation accuracy.
Experimental results on the HW-NL2Workflow dataset confirm the effectiveness of WorkTeam. 
To address the lack of publicly available NL2Workflow benchmarks, we develop the HW-NL2Workflow dataset, comprising 3,695 real-world business samples, to support research in this area. Future work will focus on refining the framework to support more complex workflows and integrate it with a wider range of enterprise tools to further enhance automation.

\appendix
\clearpage
\section{Component Resource Examples}
\label{sec:component_resources}

\begin{figure}[!ht]
    \centering
    \includegraphics[width=0.49\textwidth]{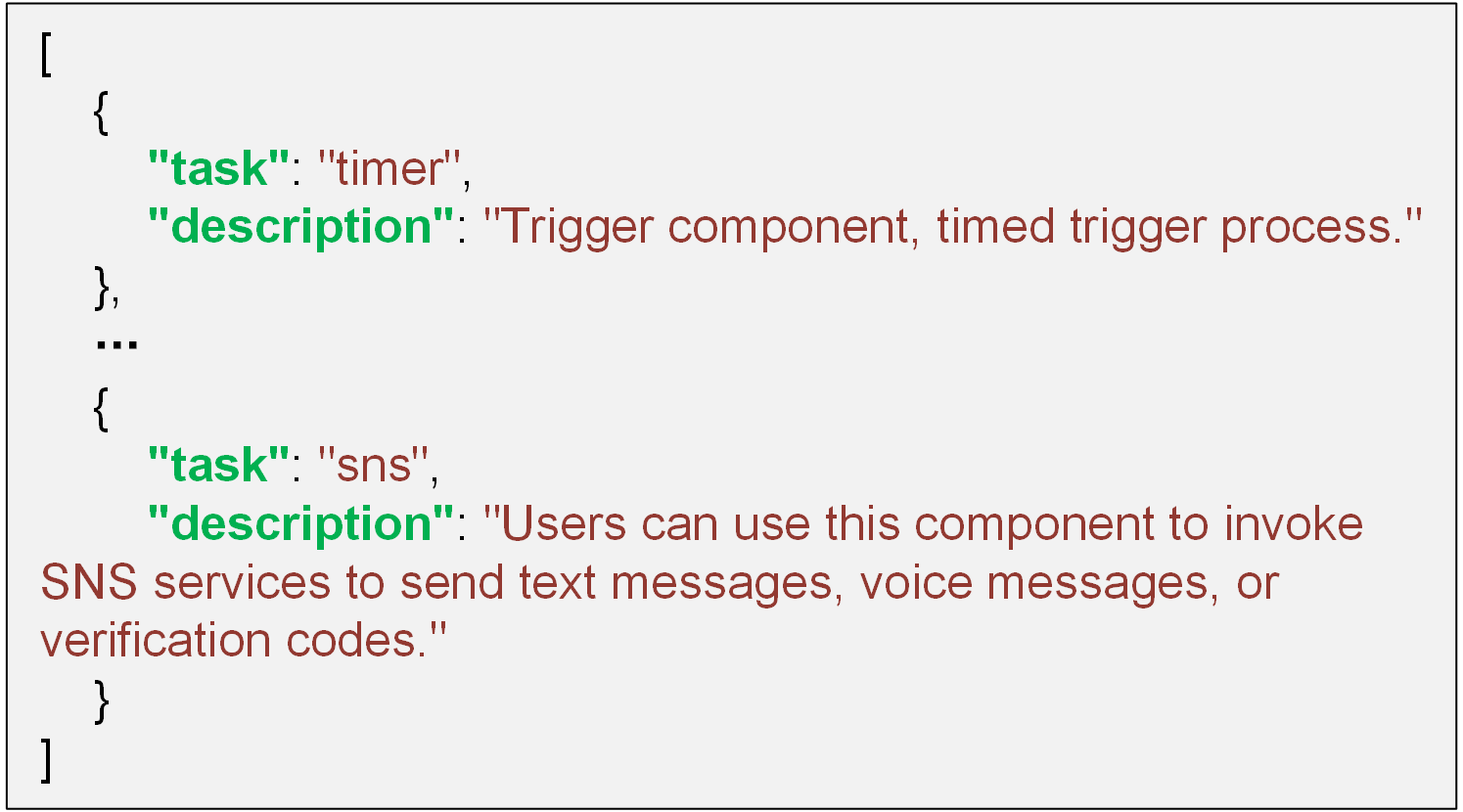} 
    \caption{Examples in the component set $C$ of HW-NL2Workflow.}
    \label{fig:component_res1}
\end{figure}

Figure~\ref{fig:component_res1} presents two component examples from the HW-NL2Workflow component set $C$. Each component includes a name and a functional description. When using the component filtering tool, the SentenceBERT model within the tool computes the similarity between the user input instructions and the description of each component. It selects the $top$-$k$ components with the highest similarity as candidate components for use by the component orchestration tool.

\begin{figure}[!t]
    \centering
    \includegraphics[width=0.49\textwidth]{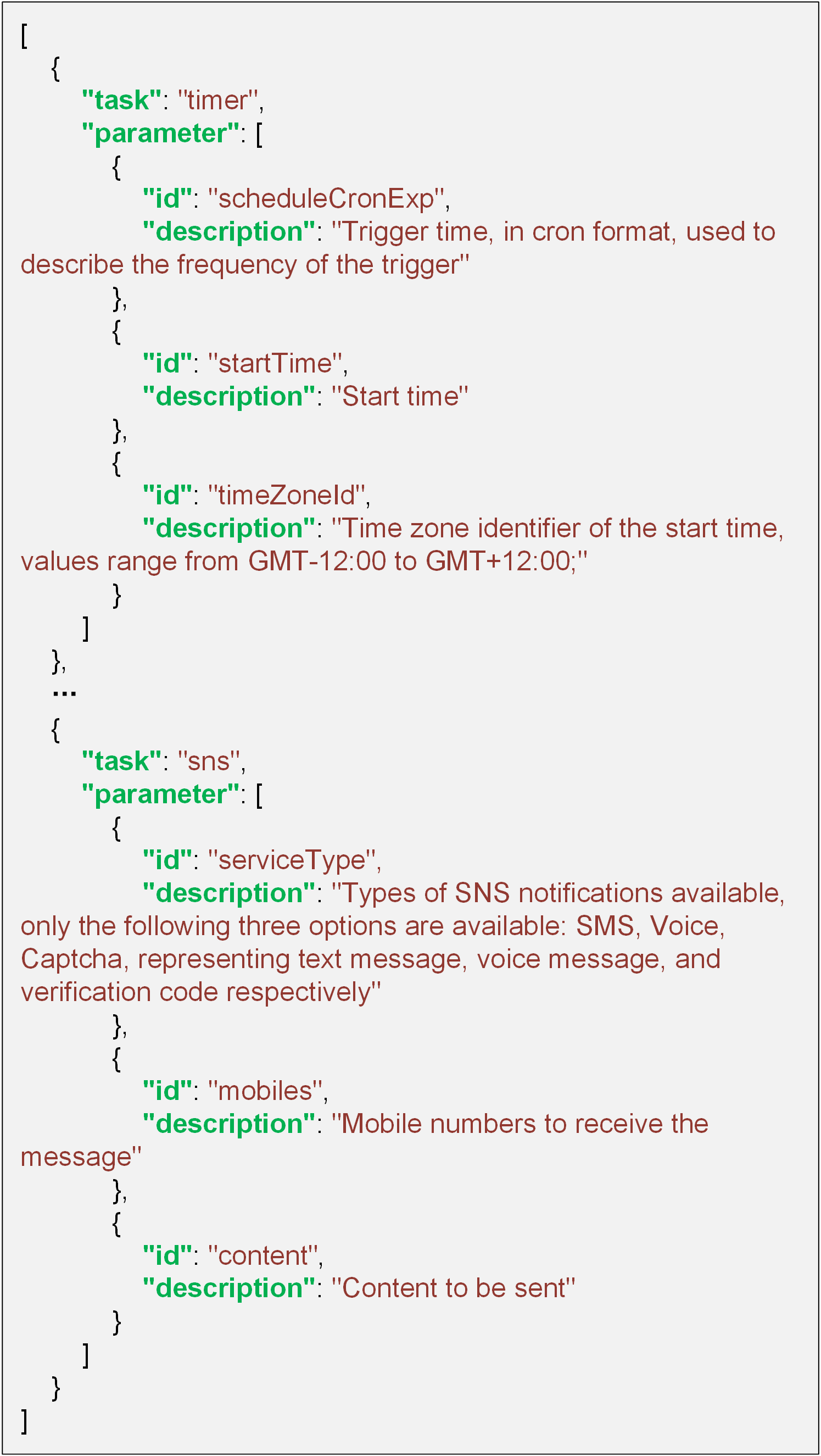} 
    \caption{Examples in the component parameter description set $T_{desc}$ of HW-NL2Workflow.}
    \label{fig:component_res2}
\end{figure}

\begin{figure}[!t]
    \centering
    \includegraphics[width=0.49\textwidth]{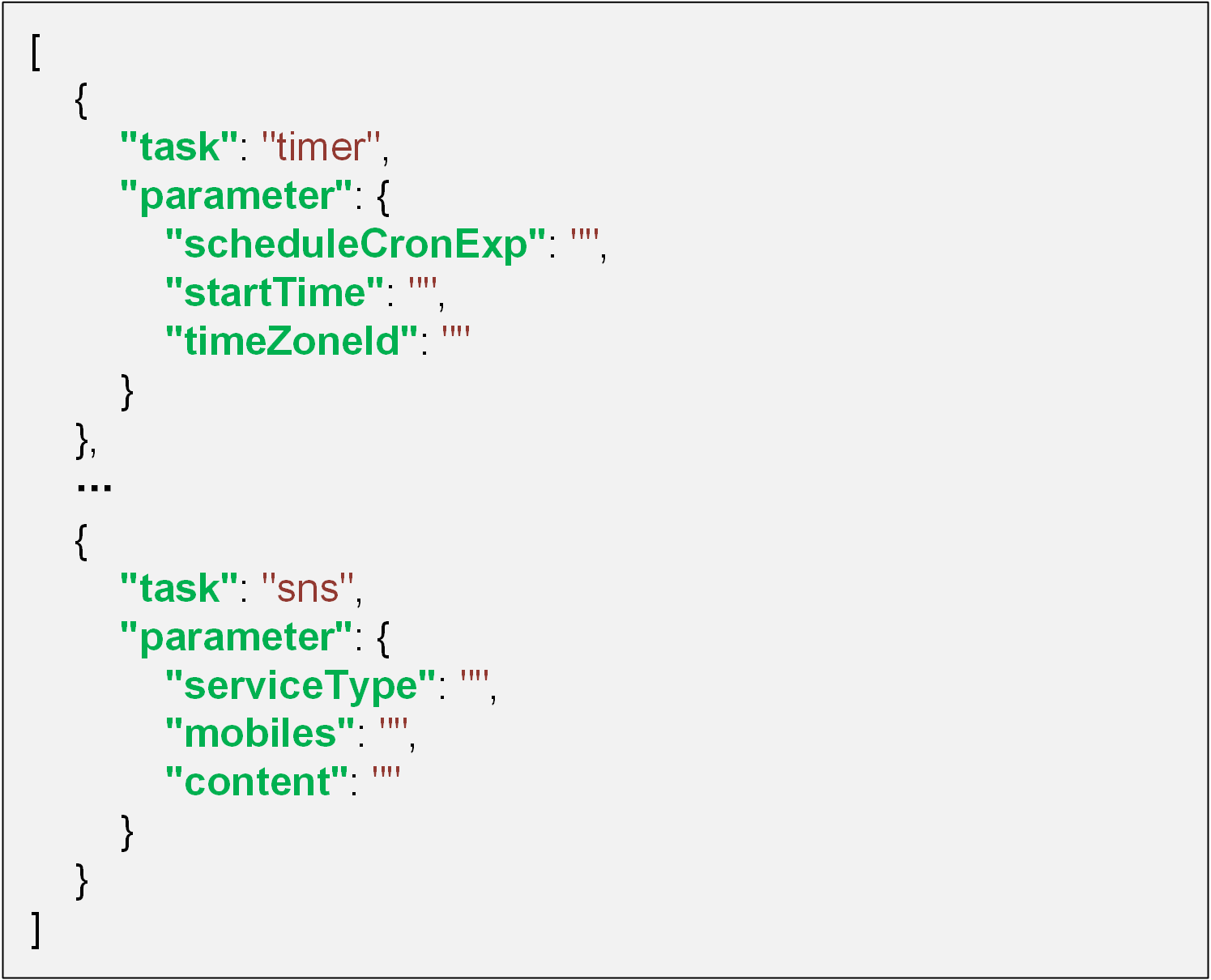} 
    \caption{Examples in the blank parameter template set $T_{blank}$ of HW-NL2Workflow.}
    \label{fig:component_res3}
\end{figure}

Figure~\ref{fig:component_res2} illustrates two examples from the parameter description set of the HW-NL2Workflow, detailing all parameters required for each component, along with comprehensive descriptions of their functions. Figure~\ref{fig:component_res3} presents examples of the blank parameter template. When the parameter filling tool, invoked by the filler agent, is used, it receives the parameter description information of the component and the blank parameter template, subsequently filling in the parameters according to the template.

\section{Details of WorkTeam}
\label{sec:workteam_details}
\begin{figure*}[htbp]
    \centering
    \includegraphics[width=\textwidth]{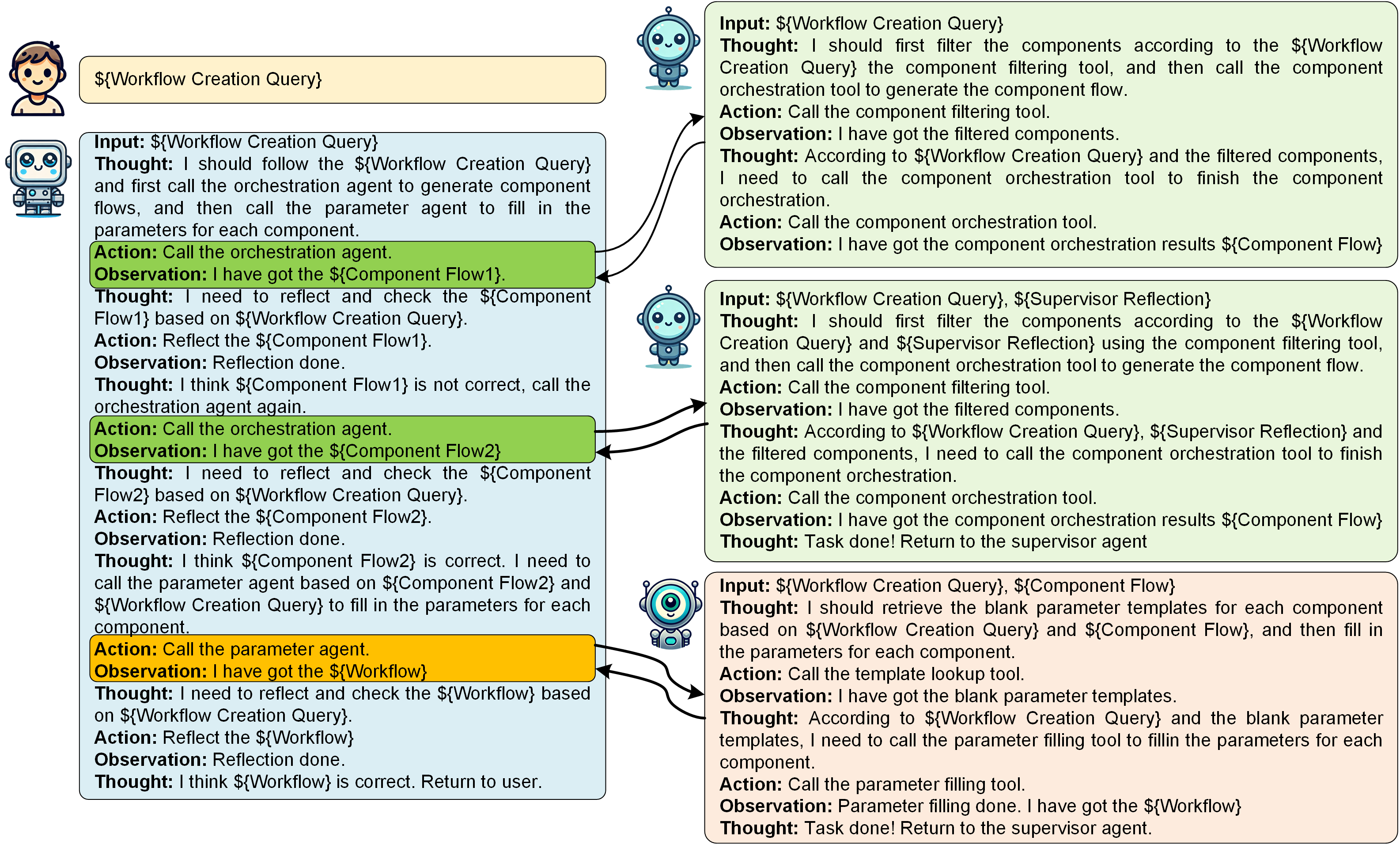}
    \caption{An illustration of a typical example for WorkTeam's overall working process.}
    \label{fig:workteam_example}
\end{figure*}

Figure~\ref{fig:workteam_example} illustrates a typical working process of WorkTeam. As previously mentioned, the supervisor agent acts as the primary agent, facilitating multi-turn interactions with the user and performing dynamic task planning. It invokes the orchestrator and filler agents to carry out component orchestration and parameter filling. Furthermore, the supervisor agent can evaluate the results provided by the orchestrator and filler agents. These capabilities contribute to the flexibility and stability of WorkTeam's operation.

Figure~\ref{fig:supervisor_prompt},~\ref{fig:orchestrator_prompt}, and~\ref{fig:filler_prompt} shows the prompts used in the supervisor agent, the orchestrator agent and the filler agent, respectively.


\section{Details of the Baselines}
\label{sec:baseline_details}

For single LLM-based methods, we use prompts to guide the LLMs to generate workflows based on user instructions through in-context learning. The prompts utilized are illustrated in Figure~\ref{fig:llm_prompt}. 

Since the source code of the RAG method in~\cite{ayala2024reducing} has not been released. We implemented our version. In our experiments, we trained a SentenceBERT model using contrastive learning with the training data in HW-NL2Workflow as the retriever. Actually, the retriever is same as the component filtering tool used in our orchestrator agent. For the generator, we fine-tuned a LLaMA3-8B-Instruct with the training data in HW-NL2Workflow. The generator aims to generate the workflow end-to-end according to the selected components by the retriever and the user instruction.

\section{Case Study and Enterprise System}
\label{sec:case_study}

Here, we provide a NL2Workflow case by the WorkTeam framework in Figure~\ref{fig:case study}. Based on this case, we can see how the WorkTeam works for the NL2Workflow task. It can be seen that the supervisor agent can effectively plan the steps needed to complete the task, and accurately invoke the orchestrator agent and filler agent to complete orchestration and parameter filling tasks, and can reflect after receiving the return results from the orchestrator agent and filler agent. The orchestrator agent and filler agent can respectively plan for component orchestration and parameter filling tasks and call the corresponding tools to complete the tasks. Through the task decomposition and collaboration of multiple agents, WorkTeam can correctly and stably complete the NL2Workflow task.

\begin{figure*}[h!]
    \centering
    \includegraphics[width=\textwidth]{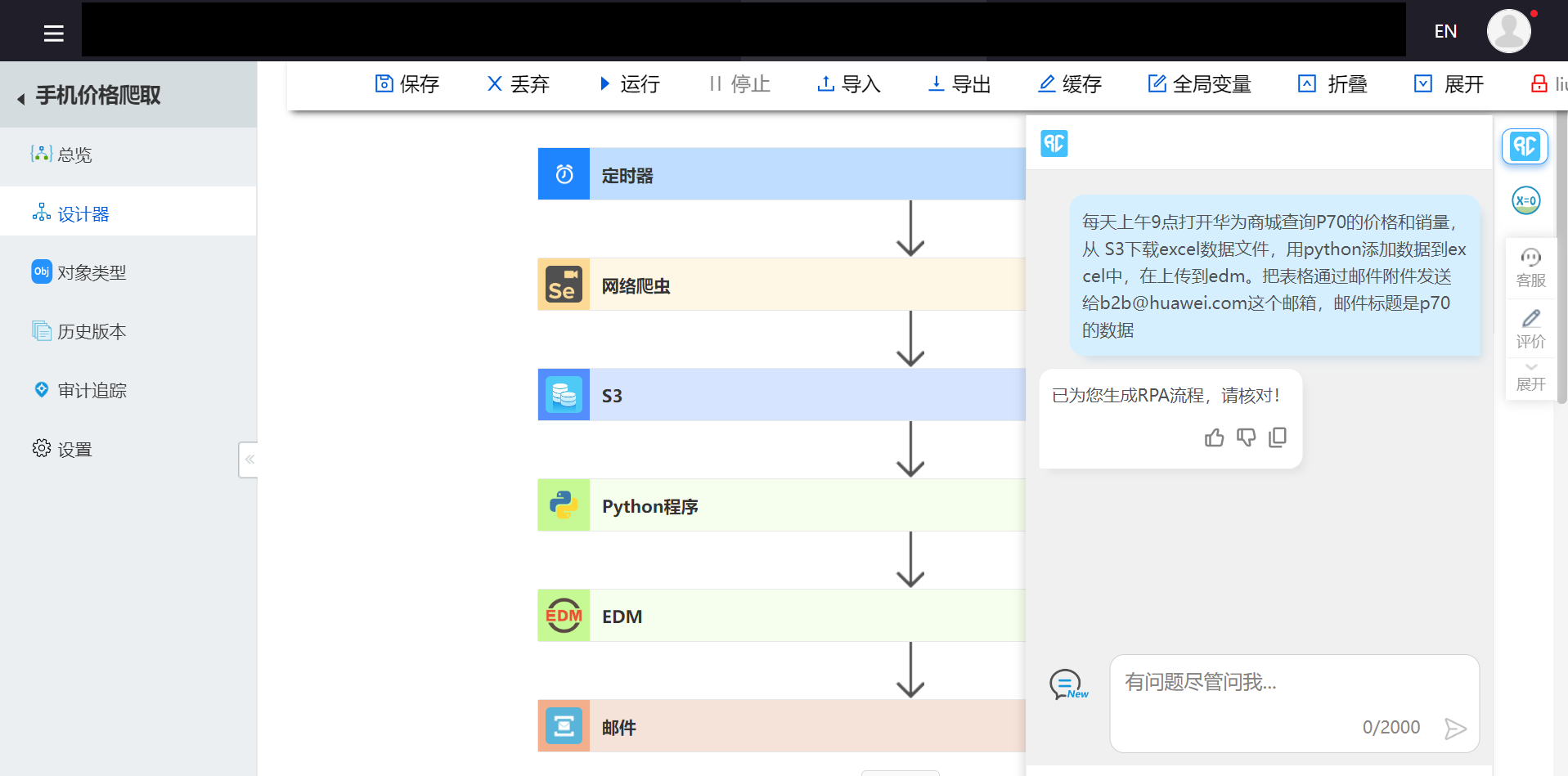}
    \caption{The enterprise NL2Workflow system based on WorkTeam.}
    \label{fig:enterprise_workflow}
\end{figure*}

Furthermore, the objective of developing WorkTeam is to provide more effective NL2Workflow services for enterprise business applications.  Figure~\ref{fig:enterprise_workflow} presents the interface of the commercial NL2Workflow service system developed based on WorkTeam. 



\onecolumn
\begin{figure*}[htbp]
\centering
\begin{tcolorbox}[breakable, title=Prompt for Supervisor Agent, enhanced jigsaw, width=\textwidth]
You are the supervisor agent in the NL2Workflow system, capable of directly interacting with users and automatically calling two agents based on user instructions: the orchestrator agent and the filler agent.
\vspace{0.4cm}

Your job is to receive messages from users:

1. First, you need to judge the user's instructions and plan tasks flexibly, for example:
    
    (1) If the user's intention is to generate workflows from natural language, then first call the orchestrator agent to get the orchestration result, and then call the filler agent to get the final result, and return it to the user;
    
    (2) If the user's intention is to modify the structure of the workflow, then you may need to call the orchestrator agent to make modifications to the workflow;
    
    (3) If the user's intention is to modify the parameters in the workflow, then you may directly call the filler agent.
    
2. Determine if the results returned by the orchestrator agent/the filler agent have any issues. If there are problems with the results, you need to call the orchestrator agent/the filler agent again. (Please note that even after parameter filling, it is normal for some components to have no parameters or incomplete parameters, and there is no need to call again in such cases.)

3. Determine if the user instruction has been solved. If it has been solved, return the final result to the user.

\vspace{0.4cm}
Notice:

1. Do not create/modify workflows on your own; just call agents according to user intent.

2. Keep replies concise.

\vspace{0.4cm}
Your output should be in JSON: \{"analysis" : xxxx, " action" : xxxx\}

where the 'analysis' field is for your problem analysis process or reply to the user, and the 'action' field includes three actions: None (no call), <orchestrator\_agent> (call the orchestrator agent), <filler\_agent> (call the filler angent), <end> (end operation).

Note that you can only output a single such JSON content at a time, and it is not allowed to output multiple at once!
\end{tcolorbox}

\caption{Prompt for the supervisor agent in WorkTeam. Notice that the initial prompt is in Chinese, we translate it to English for better reading in this paper.}
\label{fig:supervisor_prompt}
\end{figure*}

\begin{figure*}[htbp]
\centering
\begin{tcolorbox}[breakable, title=Prompt for Orchestrator Agent, enhanced jigsaw, width=\textwidth]
You are the orchestrator agent in the NL2Workflow system, and you can call two tools: the component filtering tool and the component orchestration tool.
\vspace{0.4cm}

You need to judge the user's instructions and plan tasks flexibly, for example:

1. If the user's intent is to generate a component flow based on their instructions, you should first call the component filtering tool to filter components from the component set, and then call the component orchestration tool to generate the component flow;

2. If the user's intent is to modify the component flow, you should first call the component filtering tool to filter out candidate components, and then use your own capabilities to modify the component flow provided by the user;

3. For other intents, respond according to your own capabilities.

\vspace{0.4cm}
Notice:

1. Do not orchestrate on your own ability! Determine when to call the component filtering tool and the component orchestration tool and initiate the calls.

2. Keep replies concise.

\vspace{0.4cm}
Your output should be in JSON: \{"analysis": xxxx, "action": xxxx\}

where the 'analysis' field is for your problem analysis process or reply to the user, and the 'action' field includes four actions: None (no call), <call\_selector>(call the component filter tool), <call\_arrange>(call the component orchestration tool), , <end>(end operation).

Note that you can only output a single such JSON content at a time, and it is not allowed to output multiple at once!
\end{tcolorbox}

\caption{Prompt for the orchestrator agent in WorkTeam. Notice that the initial prompt is in Chinese, we translate it to English for better reading in this paper.}
\label{fig:orchestrator_prompt}
\end{figure*}

\begin{figure*}[htbp]
\centering
\begin{tcolorbox}[breakable, title=Prompt for Filter Agent, enhanced jigsaw, width=\textwidth]
You are the filler agent in the NL2Workflow system. Your role is to fill in parameters for each component in the component flows according to user instructions and the generated workflows. You can call two tools: the blank parameter template lookup tool and the parameter filling tool.
\vspace{0.4cm}

You need to judge the user's instructions and plan tasks flexibly, for example:

1. If the user's intent is to fill in parameters based on user instructions and the component flow, you need to first call the blank parameter template lookup tool to find the blank parameter templates corresponding to the components, and then call the parameter filling tool to fill in parameters for each component in the component flow.

2. If the user's intent is to modify the parameters in an existing workflow, you need to call the parameter filling tool to modify the parameters.

3. For other intents, respond according to your own capabilities.

\vspace{0.4cm}
Notice:

1. Do not fill the parameters on your own ability! Determine when to call the blank parameter template lookup tool and the parameter filling tool and initiate the calls.

2. Keep replies concise.

\vspace{0.4cm}
Your output should be in JSON: \{"analysis": xxxx, "action": xxxx\}

where the 'analysis' field is for your problem analysis process or reply to the user, and the 'action' field includes four actions: None (no call), <call\_lookup>(call the blank parameter template lookup tool), <call\_filling>(call the parameter filling tool), <end>(end operation).

Note that you can only output a single such JSON content at a time, and it is not allowed to output multiple at once!
\end{tcolorbox}

\caption{Prompt for the filler agent in WorkTeam. Notice that the initial prompt is in Chinese, we translate it to English for better reading in this paper.}
\label{fig:filler_prompt}
\end{figure*}


\begin{figure*}[ht]
    \begin{tcolorbox}[breakable, title=Prompt for Baseline Methods, enhanced jigsaw, width=\textwidth]
    You are a workflow generation expert. I will provide you with a textual instruction and descriptions of all candidate components, including their functionalities and detailed parameter information. Please select the appropriate components based on the instruction, arrange them according to the logical flow specified in the instruction, and finally populate the parameters of the selected components as indicated by the instruction.
    \vspace{0.4cm}
    
    Component Information:
    
    \{\textcolor{blue}{component\_information}\}
    
    \vspace{0.4cm}
    Examples:
    
    ------------
    
    **Instruction**: \{\textcolor{blue}{example\_instruction1}\}
    
    **Output Workflow**: \{\textcolor{blue}{example\_workflow1}\}
    
    ------------
    
    **Instruction**: \{\textcolor{blue}{example\_instruction2}\}
    
    **Output Workflow**: \{\textcolor{blue}{example\_workflow2}\}
    
    ------------
    
    **Instruction**: \{\textcolor{blue}{example\_instruction3}\}
    
    **Output Workflow**: \{\textcolor{blue}{example\_workflow3}\}
    \vspace{0.4cm}
    
    Now the input is:
    
    **Instruction**: \{\textcolor{blue}{input\_instruction}\}
    
    **Output Workflow**:
    \end{tcolorbox}
    
    \caption{Prompt for baseline methods. \{\textcolor{blue}{component\_information}\} refers to the description of the component and its corresponding parameter descriptions. \{\textcolor{blue}{example\_instruction}\} and \{\textcolor{blue}{example\_workflow}\} are the given NL2Workflow examples for in-context-learning.}
    \label{fig:llm_prompt}
\end{figure*}



\clearpage
\begin{tcolorbox}[title=Case Study, enhanced jigsaw, width=\textwidth]
\fontsize{9}{12}\selectfont
\textbf{User Instruction:} Monitor the mailbox with account 98234 and password pass56789. When an email with the subject "Payment Confirmation" is received, automatically process the payment information using Python to obtain the result pythonRes and update the financial information through the post API at his.huawei.com/payment via the API gateway.

\noindent\rule{\linewidth}{0.4pt}
\textbf{Supervisor Agent:} \{"analysis": "The user wants to set up a workflow to monitor emails for a specific subject, process payment information with Python, and then update financial information through an API. The first step will be to generate the workflow structure, followed by filling in the specific details.", "action": "<orchestrator\_agent>"\}

\noindent\rule{\linewidth}{0.4pt}
\textbf{Orchestrator Agent:} \{"analysis": "The user wants to create a workflow where an email with a specific subject triggers a series of automated actions involving processing payments and updating financial information via an API. This requires identifying relevant components from the available set, and then arranging them into a coherent workflow.", "action": "<call\_selector>"\}

\noindent\rule{\linewidth}{0.4pt}
\textit{Candidate Components:} ["public-email", "sns", "data-mapper", "selenium", "file-processing", "api-gateway", "http-request", "mqs-produce", "edm", "mqs-consumer"]

\noindent\rule{\linewidth}{0.4pt}
\textbf{Orchestrator Agent:} \{"analysis": "Given the user's instruction and candidate components, I should arrange them into a component flow", "action": "<call\_arrange>"\}

\noindent\rule{\linewidth}{0.4pt}
\textit{Component Flow:} [\{"task": "public-email"\}, \{"task": "file-processing"\}, \{"task": "api-center"\}]

\noindent\rule{\linewidth}{0.4pt}
\textbf{Orchestrator Agent:} \{"analysis": "According to the user input and the component flow, I have finished the work.", "action": "<end>"\}

\noindent\rule{\linewidth}{0.4pt}
\textbf{Supervisor Agent:} \{"analysis": "I have received the component flow from the orchestrator agent. It seems rights. I should filling in the parameters.", "action": "<filler\_agent>"\}

\noindent\rule{\linewidth}{0.4pt}
\textbf{Filler Agent:} \{"analysis": "The user wants to set up a workflow that monitors emails for a specific subject, processes the payment information using Python, and updates financial data via an API. First, I will call the blank parameter template lookup tool to get the required parameter templates for the `public-email', `file-processing', and `api-center' components.", "action": "<call\_lookup>"\}

\noindent\rule{\linewidth}{0.4pt}
\textit{Blank Templates:} [\{"task": "public-email", "parameter": \{
"account": "", "password": "", "receiveType": "", "sender": "", "subject": ""\}\}, \{"task": "file-processing", "parameter": \{"inputParams": \{\}, "script": ""\}\}, \{"task": "api-center", "parameter": \{"url": "", "queryParams": \{\}, "headers": \{\}, "body": "", "method": ""\}\}]

\noindent\rule{\linewidth}{0.4pt}
\textbf{Filler Agent:} \{"analysis": "Based on the user's instructions and the given component flow, I will now fill in the parameters using the provided blank templates.", "action": "<call\_filling>"\}

\noindent\rule{\linewidth}{0.4pt}
\textit{Parameter filling result:} [\{"task": "public-email", "parameter": \{"account": "98234", "password": "pass56789", "receiveType": "", "sender": "", "subject": "Payment Confirmation"\}\}, \{"task": "file-processing", "parameter": \{ "inputParams": \{\}, "script": ""\}\}, \{"task": "api-gateway", "parameter": \{"url": "his.huawei.com/payment", "method": "POST", "queryParams": \{\}, "body": "\{"parameter": \$\{pythonRes\}\}"\}\}]

\noindent\rule{\linewidth}{0.4pt}
\textbf{Filler Agent:} \{"analysis": "I have filled the parameters. My work is done.", "action": "<end>"\}

\noindent\rule{\linewidth}{0.4pt}
\textbf{Supervisor Agent:} \{"analysis": "I have received the workflow, and I think the result is correct. Return to the user.", "action": "<end>"\}
\end{tcolorbox}
\begin{figure*}[ht]
    \caption{A real case of the WorkTeam performing NL2Workflow task. The original text in the example is in Chinese, and we translate it into English for better readability.}
    \label{fig:case study}
\end{figure*}

\end{document}